\pdfoutput=1
\documentclass[10pt,twocolumn,letterpaper]{article}

\usepackage{wacv}
\usepackage{times}
\usepackage{epsfig}
\usepackage{graphicx}
\usepackage{amsmath}
\usepackage{amssymb}

\usepackage{algorithmic}
\usepackage{textcomp}
\usepackage{xcolor}
\usepackage{multirow}
\usepackage{color}
\usepackage{subfigure}
\usepackage{cite}

\usepackage{caption}
\usepackage{adjustbox}
\usepackage{wrapfig}

\def\wacvPaperID{0045} 

\wacvfinalcopy 

\ifwacvfinal
\def\assignedStartPage{9876} 
\fi

\ifwacvfinal
\usepackage[breaklinks=true,bookmarks=false]{hyperref}
\else
\usepackage[pagebackref=true,breaklinks=true,colorlinks,bookmarks=false]{hyperref}
\fi

\ifwacvfinal
\else
\fi

\begin{document}

\title{BiHPF: Bilateral High-Pass Filters for Robust Deepfake Detection}
\author{Yonghyun Jeong$^1$, Doyeon Kim$^1$, Seungjai Min$^1$, Seongho Joe$^1$, Youngjune Gwon$^1$, Jongwon Choi$^2$\thanks{Corresponding author.}\\
$^1$Samsung SDS, Seoul, Korea\\
{\tt\small  yhyun.jeong, dy31.kim, seungjai.min, drizzle.cho,  gyj.gwon@samsung.com}
\and
$^2$Dept. of Advanced Imaging, Chung-Ang University, Seoul, Korea\\
{\tt\small choijw@cau.ac.kr}
}

\maketitle

\begin{abstract}
The advancement in numerous generative models has a two-fold effect: a simple and easy generation of realistic synthesized images, but also an increased risk of malicious abuse of those images. Thus, it is important to develop a generalized detector for synthesized images of any GAN model or object category, including those unseen during the training phase. However, the conventional methods heavily depend on the training settings, which cause a dramatic decline in performance when tested with unknown domains. To resolve the issue and obtain a generalized detection ability, we propose \textit{Bilateral High-Pass Filters (BiHPF)}, which amplify the effect of the frequency-level artifacts that are known to be found in the synthesized images of generative models. Numerous experimental results validate that our method outperforms other state-of-the-art methods, even when tested with unseen domains. 
\end{abstract}
\section{Introduction}
Recently, the advancement of generative models including Generative Adversarial Networks (GAN)~\cite{stylegan2} has allowed the easy generation of realistic synthesized images. 
Unfortunately, these generated images can be abused with malicious purposes and ultimately bring detrimental consequences in societal, political, and economic aspects, such as fraud, defamation, and fake news~\cite{lee,kwon,huh,nguyen2019deep,dofnet}. 
Thus, it is highly important to develop a robust detection model to protect our society~\cite{sun,dofnet}.

\begin{figure}[t] 
\centering
\includegraphics[width=0.85\linewidth]{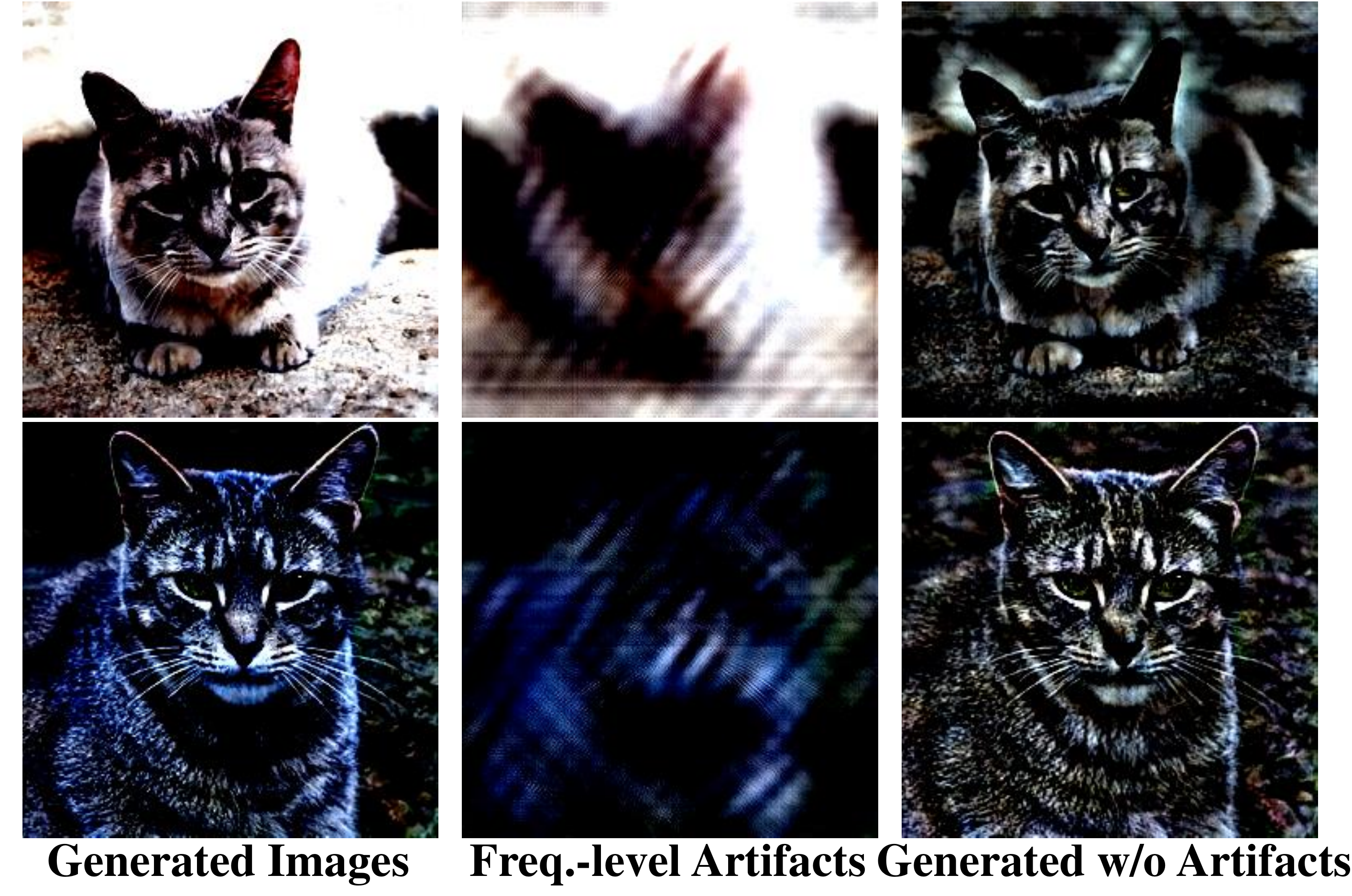}
\caption{\textbf{Visualization of the frequency-level artifacts discovered in synthesized images.} The artifacts are extracted by the proposed classification network adversarially trained for a frequency-level compression map. Based on the analysis of the averaged artifacts, we develop a novel mechanism using Bilateral High-Pass Filters (BiHPF), which improve cross-domain performance for detecting the synthesized images.} 
\label{fig:teasor} 
\end{figure}

A number of studies have been conducted for detecting the generated images by GAN models. 
Various studies including \cite{adobe,watch_cvpr20,frank,unmasking} have tackled the entire image synthesis of diverse categories, while others focus on manipulations on the human face only~\cite{headpose,eyeblinking,agarwal_protecting_2019,matern,montserrat,dang,facexray}.
However, most of the prior literature suffers from a dramatic decline in performance when tested with unknown domains outside of the training data, due to their domain-specific detection.
Since it is easy to switch the target domain of the generated images with malicious purposes, it is important to achieve robust detection even in unseen domains~\cite{sun}.
This capability can be defined as \textit{cross-domain performance}.
In this paper, the domains include various properties, such as the categories of the generated subjects, the color manipulations, and GAN models.
As the range of such domains rapidly expands with technological advancement, we believe it is more important to focus our attention to the generalized detection of the synthesized images by GAN models, rather than focusing on the facial manipulations only. 

It has been confirmed by several previous studies~\cite{zhang,adobe,frank} that the synthetic images of generative models contain unique artifacts caused by the upsampling of GAN pipeline. 
We find that these artifacts in the frequency spectrum can become a key factor in developing a robust detector. 
Thus, we develop an adversarial framework to train the `artifact compression map,' which is multiplied into the frequency spectrum to reduce the artifacts.
As shown in the second column of Fig.~\ref{fig:teasor}, the intensity of the artifact compression map represents the degree of artifacts contained in the frequency spectrum.
By analyzing the trained compression map, we obtain two practical insights. 
First, the artifacts have large magnitudes in the high-frequency components, as discovered in our numerous experiments in the frequency spectrum. 
Second, the artifacts are located in the surrounding background of the image rather than the central region, as observed in the pixel-level image transformed from the frequency-level compression map.
We notice the two important discoveries are generally observed in various settings, which provide an insight that utilizing the artifacts can prevent domain-specific detection and enhance cross-domain detecting performance.

Thus, we design a simple mechanism called Bilateral High-Pass Filters (BiHPF) to emphasize the effects of the artifacts for robust detection.
BiHPF consists of two High-Pass Filters (HPF): the frequency-level HPF for amplifying the magnitudes of the artifacts in the high-frequency components, and the pixel-level HPF for emphasizing the pixel values in the surrounding background in the pixel domain. 
Then, the classification model utilizes the BiHPF-processed magnitude spectrum map as the input of the network.
To validate the overall performance of the proposed algorithm, we consider various domains including the unseen categories of diverse subjects, color settings, and GAN models.
Numerous experiments confirm that our proposed algorithm significantly improves the detecting performance.

\section{Related Work}
The detection methods for CNN-based manipulations and synthesized images by GAN models can be classified into two major categories based on the input data: pixel-based detection and frequency-based detection.\\
\textbf{Pixel-based detection. }   
Some of the previous studies target image pixels as the input data for the detection of image forgery. The earlier studies were focused on specific conditions, such as compression and lighting. 
Some analyzed inconsistencies in blocking artifacts generated during JPEG compression~\cite{ye, tralic}, while others focused on 3D lighting and assessed inconsistencies in lighting conditions of subjects in images to detect manipulations~\cite{kee, carvalho, peng, peng2}
To distinguish more types of tampering operations,~\cite{dirik, ferrara} suggested analyzing the demosaicing artifacts generated by Color Filter Array (CFA) processing in tampered images; however, this method was inapplicable to resized images, since the artifacts disappeared during the resizing process. 

Recently, with the rise of deepfakes, most studies focused on the temporal properties, such as facial features~\cite{agarwal_protecting_2019,matern,resnet,montserrat}, incoherent head poses~\cite{headpose}, and lack of eye-blinking~\cite{eyeblinking}, and various studies~\cite{faceforensics++, DFDC, Celeb_DF_cvpr20} provided large-scale datasets and evaluated various methods of image forensics for face manipulations. 
To broaden the scope of detection,~\cite{bayar} designed a specific convolutional layer to learn prediction error filters for a generalized detection model; however, it shows a decline in performance 
when numerous post-processing methods were employed to manipulated regions.
Thus, recent studies have worked on developing a generalized model with improved performance as in~\cite{cozzolino}, which introduced an adaptable autoencoder-based neural network architecture to new target domains using a few training samples. 
Also,~\cite{adobe} employed RGB images as training inputs for the classifiers to distinguish cross-model manipulations by post-processing operations as blurring and JPEG compression.\\
\textbf{Frequency-based detection. }  
Some other studies employ frequency spectrum as training input for the classifier to distinguish between the real and fake images. Previous studies, such as \cite{kirchner}, suggested a detection method based on the artifacts in the spatial, frequency domain through the variance of the prediction residue. 
To enlarge the detection scope, \cite{huang} employed FFT and SVD to distinguish copy-move manipulations in images under JPEG compression and Gaussian noise and blurring attacks. Also, \cite{marra} suggested a GAN-specific detection method in frequency-domain based on artificial fingerprints on generated images.
Another study \cite{bappy} proposed a manipulation localization architecture that utilizes spatial maps and frequency domain correlation to examine the distinct features of forged areas by employing an encoder and LSTM network. 
Recently, \cite{frank} conducted a comprehensive analysis of the artifacts generated by GANs in the frequency space using Discrete Cosine Transfer (DCT). Also, \cite{zhang} proposed a classifier model based on the artifacts induced by the up-sampler of GANs, however it exploited numerous categories of datasets, including the face, horse, landscape, satellite image, and painting. \cite{watch_cvpr20, unmasking} exploited the spectral distortions via azimuthal integration for identifying inauthentic images. 

Our approach also employs frequency spectrum as the input data but differs from the others in two major aspects: first, we employ the high pass filter at the low-frequency spectrum to solely concentrate on the high-frequency spectrum; second, Laplacian of Gaussian (LoG) is applied to the magnitude for our model to focus on the background of the image to identify the general characteristics of synthesized images for cross-domain detection.
\section{Analysis of Frequency-level Artifacts}
\begin{figure*}[t]
\centering
        \includegraphics[width=0.80\linewidth]{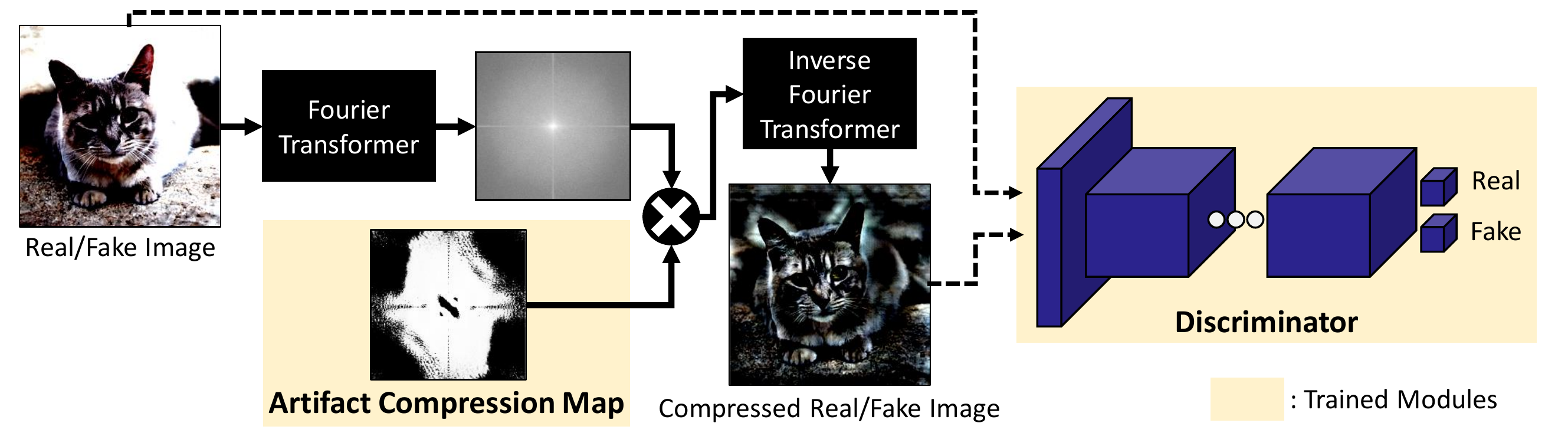}
\caption{\textbf{Network architecture for artifact compression map.} Based on the frequency-level compression map and a conventional classification network, the Artifact Compression Map (ACM) is estimated by an adversarial learning scheme.}
    \label{fig:appendix_architecture}  
\end{figure*}
Before developing a new framework to improve the cross-domain performance, we first analyze the frequency-level artifacts discovered in the synthesized images by GAN models, for the detector to distinguish between the `real images' and `fake (synthesized) images.' 
Unfortunately, since the previous methods only focus on detecting the artifacts using the labeled dataset, it has been impossible to extract and analyze the information of the frequency-level artifacts. 
Thus, we develop a novel model to extract the properties of the artifacts in the frequency spectrum, which is titled as \textit{Artifact Compression Map} (ACM).

\subsection{Artifact Compression Map}
We derive the frequency-level map of the generated images by the summation of the frequency-level contents and the frequency-level artifacts.
Thus,
\begin{equation}
    \left|\mathbf{Z}_{fake}\right| \equiv \left|\mathcal{F}\{\mathbf{X}_{fake}\}\right| = \left|\mathbf{Z}_{content}\right| + \left|\mathbf{U}\right|,\\
\end{equation}
where $|\bullet|$ results in the element-wise magnitudes of the values in the input matrix, $\mathbf{X}_{fake}$ indicates the fake image, and $\mathbf{Z}_{content}$ and $\mathbf{U}$ indicate the frequency-level content map and the frequency-level artifacts, respectively.
We also define the frequency-level real image by $\left|\mathbf{Z}_{real}\right| \equiv \left|\mathcal{F}\{\mathbf{X}_{real}\}\right|$ where $\mathbf{X}_{real}$ indicate the real image.

According to the previous studies~\cite{watch_cvpr20,unmasking}, the difference in the frequency spectrum between the real and fake images can be discovered in the specific frequency components.
We define a trainable compression map by $\mathbf{W}_c\in\mathtt{R}^{w\times h}$ where the values are one for the frequency components of the frequency-level artifacts, while all the other values are zero.
Then, the magnitude of $\mathbf{U}$ is removed by the element-wise multiplication of $\mathbf{W}_c$ for $\left|\mathbf{Z}_{fake}\right|$ as follows:
\begin{equation}
\begin{aligned}
    \mathbf{W}_c \odot \left|\mathbf{Z}_{fake}\right| 
    &= \mathbf{W}_c \odot \left|\mathbf{Z}_{content}\right|.
\end{aligned}
\end{equation}
Then, the resulting $\mathbf{W}_c \odot \left|\mathbf{Z}_{content}\right|$ becomes indistinguishable from $\left|\mathbf{Z}_{real}\right|$ weighted by $\mathbf{W}_c$ (i.e. $\mathbf{W}_c \odot \left|\mathbf{Z}_{content}\right|\approx\mathbf{W}_c \odot \left|\mathbf{Z}_{real}\right|$).
Thus, since $\mathbf{W}_c$ selectively compresses the frequency components of the artifacts, we can recognize and analyze the artifacts by reversing $\mathbf{W}_c$ to $\mathbf{1}^{w\times h}-\mathbf{W}_c$.

\subsection{Fake Image Detection Network for Adversarial Training of Compression Map}
To analyze the artifacts, we develop the fake image classification network with an add-on module adversarially estimating $\mathbf{W}_c$.
In contrast to the conventional classification network, the goal of our proposed network is to acquire $\mathbf{W}_c$, rather than the detection of fake images.
Thus, we design a novel classification network composed of the add-on module considering $\mathbf{W}_c$ as a training parameter.

As shown in Fig.~\ref{fig:appendix_architecture}, the overall architecture of our network is similar to the conventional deep neural network, and we use ResNet-50~\cite{resnet} as the basic architecture.
The classification network predicts the real and fake by using the pixel-level original images, and the add-on module synthesizes the original images by compressing the frequency components by $\mathbf{W}_c$.  
The add-on module has a simple process of operation, which contains a Fourier transformer $\mathcal{F}$, an inverse Fourier transformer $\mathcal{F}^{-1}$, and the trainable weight map $\mathbf{W}_a$. 
The resolution of $\mathbf{W}_a$ is equivalent to that of the input image~($w\times h$) and ACM~($\mathbf{W}_c$), while the number of its channels is two (i.e. $\mathbf{W}_a\in\mathtt{R}^{w\times h\times 2}$). From the two channels of $\mathbf{W}_a$, $\mathbf{W}_c$ is estimated by the softmax with the temperature scaling~\cite{temperaturesoftmax} as follows:
\begin{equation}
    \mathbf{W}_c(\omega_1, \omega_2) = \frac{e^{\left(T_f\mathbf{W}_a^1(\omega_1, \omega_2)\right)}} {{e^{\left(T_f\mathbf{W}_a^1(\omega_1, \omega_2)\right)} + e^{\left(T_f\mathbf{W}_a^2(\omega_1, \omega_2)\right)}}},
\end{equation}
where $T_f$ is a temperature scaling parameter, $\omega_1\in\{1,\dots,w\}$, $\omega_2\in\{1,\dots,h\}$, and $\mathbf{W}_a^1$ and $\mathbf{W}_a^2$ represent the first and second channel of $\mathbf{W}_a$, respectively.
We represent the overall operation of the add-on module as follows:
\begin{equation}
\begin{aligned}
    \widehat{X}(x,y) &= \\ \mathcal{F}^{-1} &\{ \mathbf{W}_c(\omega_1,\omega_2) \odot \mathcal{F} \{\mathbf{X}(x,y)\}(\omega_1,\omega_2) \}(x,y),
\end{aligned}
\end{equation}
where $x\in\{1,\dots,w\}$, $y\in\{1,\dots,h\}$, $\mathbf{X}\in\mathtt{R}^{w\times h}$ is the original input data, $\mathbf{\widehat{X}}\in\mathtt{R}^{w\times h}$ is the result from the add-on module, and $\odot$ indicates the element-wise multiplication.
Thus, in the add-on module, the compression map scales the frequency components of the input images in the frequency domain.
When the input image contains RGB color channels, one $\mathbf{W}_c$ is shared across all channels.
$\mathbf{W}_a^1$ and $\mathbf{W}_a^2$ are initialized by $w^o$ and $-w^o$, respectively, where $w^o$ is a user-defined hyperparameter.
We set $w^o$ by a large value to let the initial values of $\mathbf{W}_c$ be close to $1$.
Thus, before training, the compressed image after the add-on module is almost equivalent to the original image, because $\mathbf{W}_c$ compresses none of the frequency components.

In the training phase, the classification network is trained by the mini-batch gradient descent where every iteration contains two inner updates.
At the first update, the classification network is trained simultaneously by the original input images and the compressed images obtained from the add-on module.
With the two types of images, every weight parameter except $\mathbf{W}_a$ is updated by the classification loss $\mathcal{L}_c$ to classify the real and fake images as:
\begin{equation}
\mathcal{L}_c = \sum_{i=1}^{N_b} CE(g(\mathbf{X}_i), \mathbf{y}_i) + \sum_{i=1}^{N_b} CE(g(\widehat{\mathbf{X}}_i), \mathbf{y}_i),
\end{equation}
where $CE$ is the cross-entropy loss, $N_b$ is the size of mini-batch, $g(\mathbf{X})$ means the results of $\mathbf{X}$ predicted by the ResNet-50, $\mathbf{X}_i$ and $\mathbf{y}_i$ are the $i$-th pair of data and label in the sampled mini-batch, and $\widehat{\mathbf{X}}_i$ is the compressed image from $\mathbf{X}_i$.
Since fake image detection is a binary classification task, $\mathbf{y}_i\in\{0, 1\}$ where $0$ and $1$ represent the real and fake label, respectively.
Thus, the first term considers the predictions for the original images, while the second term tries to correctly predict the compressed images.

At the second update, only the parameters of $\mathbf{W}_a$ in the add-on module are updated.
We update $\mathbf{W}_a$ according to the definition of $\mathbf{W}_c$ that confuses the classifier to mistakenly label fake images as real, by compressing the frequency components of artifacts.
Thus, in the second step, we only consider the fake data of the sampled batch in the loss, while the labels are inverted by $0$ indicating the real image as follows:
\begin{equation}
    \mathcal{L}_{adv} = \sum_{i=1}^{N_f} CE(g(\mathbf{X}_{f(i)}), 0),
\end{equation}
where $f(i)$ is the index of pair of $i$-th fake image in the sampled batch and $N_f$ is the number of fake data in the sampled batch.

\subsection{Analysis of Compression Map and Artifacts}

\begin{table}[]
\centering
\scriptsize
\caption{Comparison of prediction accuracy by \textit{Horse}. }

\label{tab:analysis}
\begin{tabular} {cccccc}
\hline
Prediction & \textit{Horse} & \textit{Cat} & \textit{Car} & \textit{Church} & All-categories\\ 
\hline
$\mathbf{X}$ & 89.1 & \textbf{86.4} & \textbf{68.5} & \textbf{55.3} & \textbf{74.9}\\\
$\mathbf{\widehat{X}}$ & \textbf{96.0} & 71.1 & 54.4 & 52.9 & 68.6\\
\hline
\end{tabular}
\end{table}

\begin{figure}[t!]
    \centering
    \subfigure[The frequency-level compression maps. The dark regions represent the frequency components of the artifacts.]{
    \includegraphics[width=0.45\linewidth]{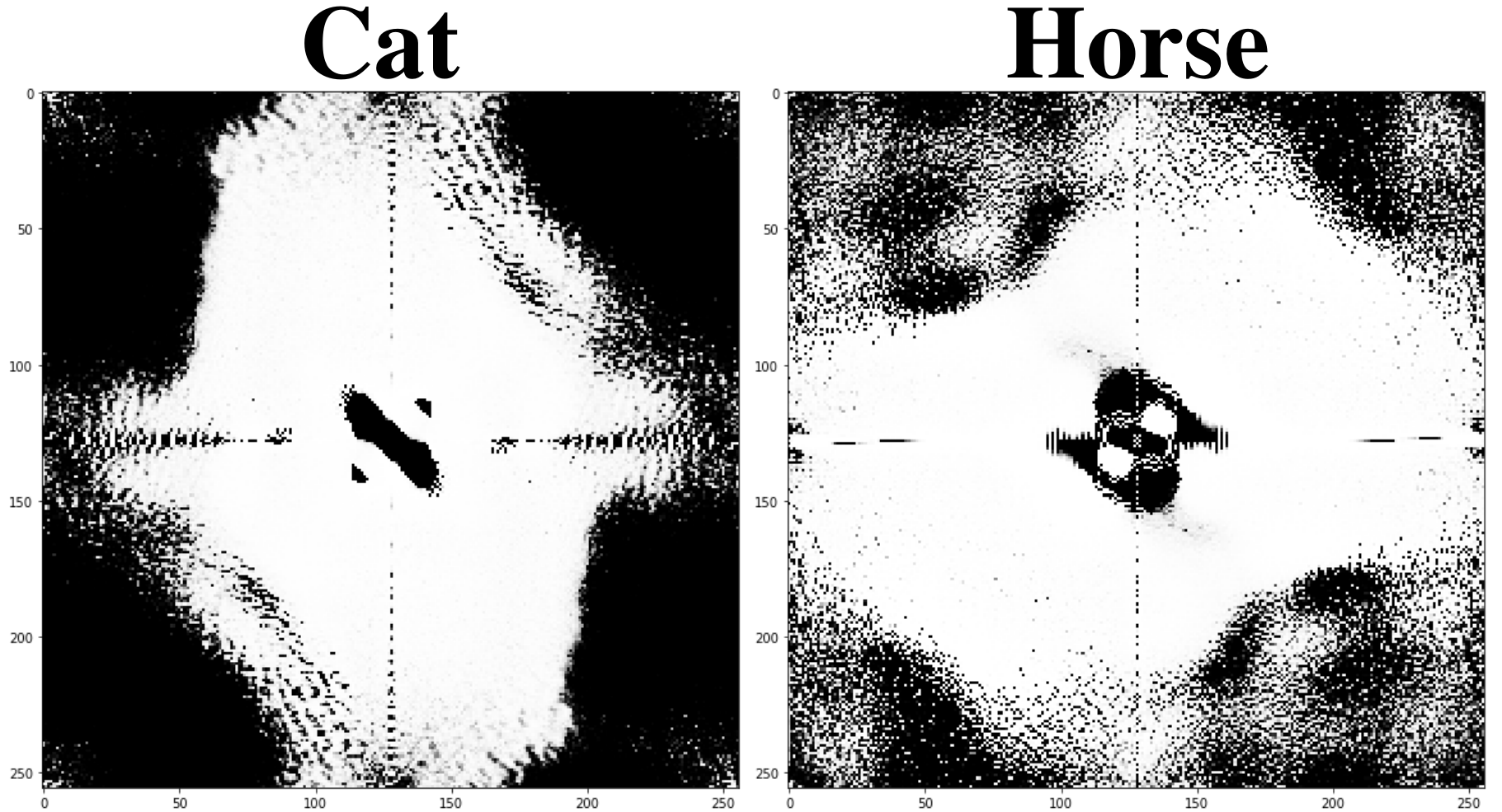}}
    \subfigure[The average maps of pixel-level artifacts. The bright pixels represent the artifacts in the background.]{
    \includegraphics[width=0.45\linewidth]{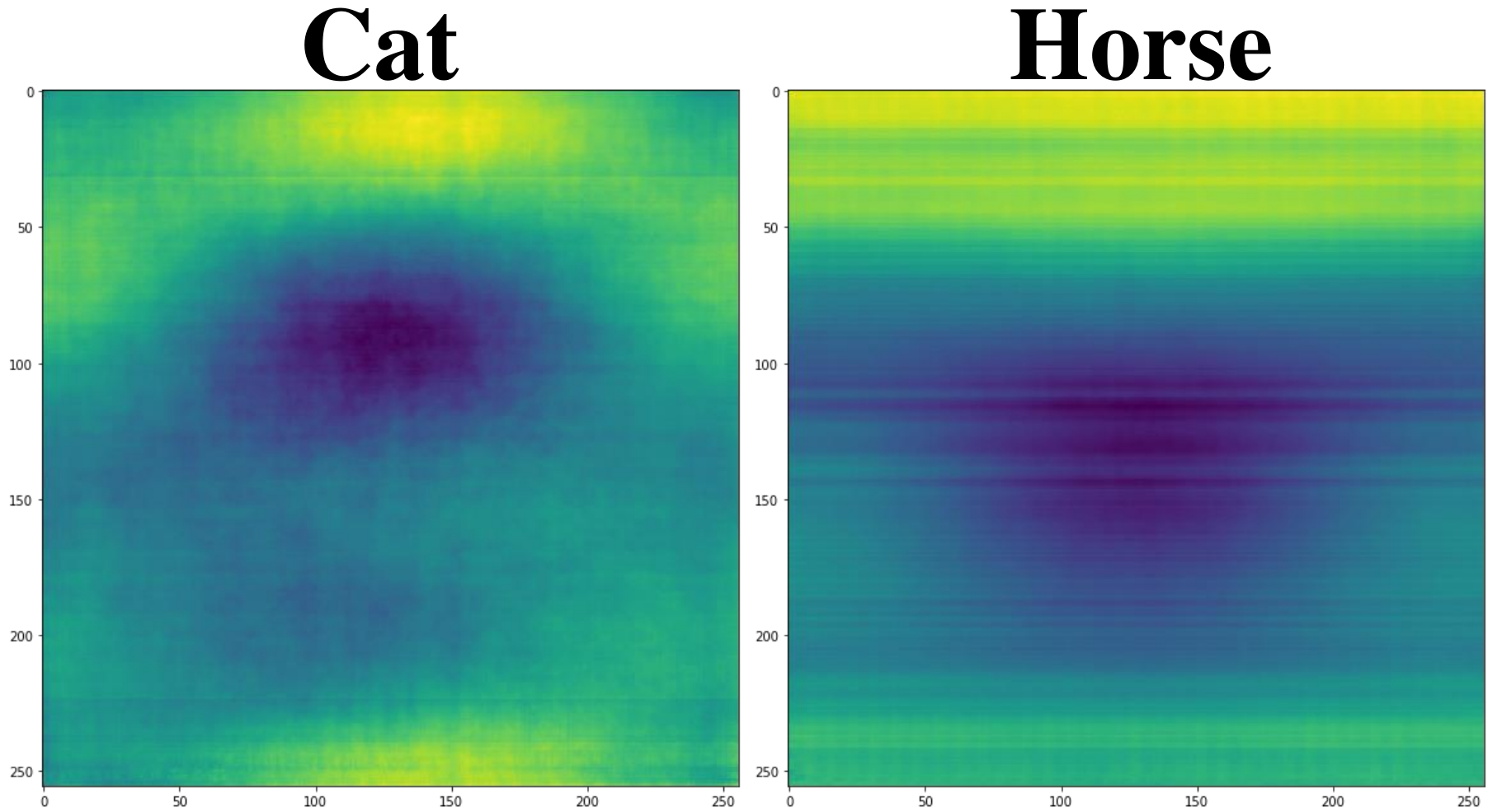}}
    \caption{\textbf{Analysis of the compression map on cat and horse.} The maps illustrate that the artifacts are found in the high frequency-level components and the pixel-level background regions.}
    \label{fig:analysis}
\end{figure}
For the analysis, we obtain the compression map by training the proposed fake image detection network with the fake datasets obtained by ProGAN~\cite{progan}.
The fake datasets contain multiple categorized fake datasets with different classes of target subjects from each other.
To show the effectiveness of the detected artifacts across the subject classes, we train the proposed network by one of the categorized datasets, which is validated by other datasets of different categories.
The datasets are explained in detail in Section~\ref{sec:implementation}.

Using the trained network, we can predict the test images by two schemes:
the first scheme is the conventional prediction method using the original images, and the second scheme is the compressed prediction method using the compressed images from the add-on module.
According to Eq.~(1), the first scheme would consider both the artifacts and the content information, while the second scheme cannot utilize the compressed artifacts.
As shown in Table~\ref{tab:analysis}, the first original scheme ($\mathbf{X}$) shows better cross-domain performance than the second compressed scheme ($\mathbf{\widehat{X}}$), which verifies that the compressed artifacts can be found across the unseen domains.
Thus, it can be confirmed that the compressed artifacts generally appear in fake images, which can be the key factor to improve the cross-domain performance.

To speculate the properties of the compressed artifacts, we analyze the trained compression map respectively by the frequency-level and pixel-level. 
Fig.~\ref{fig:analysis}(a) shows that the compression maps trained by \textit{cat} and \textit{horse} classes.
The dark regions of the compression maps represent the compressed frequency components of the artifacts.
In the trained compression map, even though the small regions of the low-frequency components in the center are also compressed, most compressed regions are located at the high-frequency components.
From the analysis, we confirm the compressed artifacts are located at the high-frequency components.

To analyze the compression map at the pixel-level, we subtract the compressed image from the original image to obtain the artifact image.
Fig.~\ref{fig:analysis}(b) shows that the average image of the artifact images obtained from the entire training images.
Interestingly, the artifacts mainly appear in the surrounding background regions in bright colors, while the central regions where subjects commonly locate appear dark.
It can be concluded that the compressed artifacts generally appear in the background region of the fake images.

\section{Bilateral High-Pass Filters}
From the trained compression map, we confirm that the artifacts mainly appear in the high-frequency components and the background region of the pixel-level images.
Based on the discoveries, we propose the Bilateral High-Pass Filters (BiHPF) that emphasize the effect of the artifacts in fake images. 
BiHPF contains two High-Pass Filters (HPF) including the \textit{pixel-level HPF} and the \textit{frequency-level HPF}.
The pixel-level HPF highlights the artifacts appearing near the backgrounds, while the frequency-level HPF emphasizes the high-frequency components where the artifacts are located to enable robust detection of fake images. 

First, the input image is transformed into the magnitude spectrum of the frequency map through the 2D Fourier transform.
If the input image has multiple channels (i.e. 3 RGB channels), we first transform the colored image into grayscale to reduce the class-specific information~\cite{matern}.
Then, we obtain the magnitude spectrum map from the frequency-level map and shift the coordinates to relocate the origin at the center. 
Here, BiHPF is applied to the magnitude spectrum map, in the consecutive order of pixel-level HPF to frequency-level HPF.
The results after applying the two HPFs are visualized in Appendix B, and the detailed operations of the HPFs are described in the following sections.
After applying HPFs, we use the filtered magnitude spectrum map as the input of the deep neural network for detection.
Our backbone architecture is based on ResNet-50~\cite{resnet}, which is pre-trained by ImageNet~\cite{imagenet}.

\subsection{Pixel-level High-pass Filter}
The pixel-level HPF works to compress the central regions of the image in the pixel-domain and to emphasize the effect of the artifacts in the background regions.
Although various options are available to compress the central region, we propose a method utilizing the frequency-level Laplacian of Gaussian (LoG) filter.
Since the LoG filter is applied to the frequency domain, the artifacts in the frequency-level are effectively focused compared to the method employing a weighting window in the pixel-level map.
Also, the LoG filter requires only one tuning hyperparameter as the variance of LoG kernel, which simplifies the tuning sequence.
To verify the equivalency of the LoG filter in the frequency magnitude spectrum and the weighting window in the pixel image, we need to utilize several properties of the Fourier transform.
For a simple description, we first consider the 1D LoG kernel.
Through the properties of differentiation and linearity of the Fourier transform, the LoG kernel with the various $\sigma^2$ can be transformed as follows: 
\begin{equation}
\begin{aligned}
    \mathcal{F}\{LoG(x)\}(\omega) &=\mathcal{F}\left\{-\frac{1}{\sigma^2}\left(1-\frac{x^2}{\sigma^2}\right)e^{-\frac{x^2}{2\sigma^2}}\right\}(\omega) \\&=-\sigma \omega^2e^{-\frac{(\sigma\omega)^2}{2}}.
\end{aligned}
\end{equation}
Then, according to the dual properties of the Fourier transform, we can estimate the inverse Fourier transform of $LoG(\omega)$ as follows:
\begin{equation}
    \mathcal{F}^{-1}\{LoG(\omega)\}(x)=-\frac{\sigma x^2}{2\pi} e^{-\frac{(\sigma x)^2}{2}}.
\end{equation}

From the derivation, the weighting window is represented as $\mathcal{F}^{-1}\{LoG(\omega)\}(x)$, which amplifies the specific regions determined by $\sigma$ while compressing the other regions.
Thus, for a 1D vector of $v$, the LoG filter in the frequency spectrum map is equivalent to the element-wise multiplication of the weighting window in the pixel domain by $\mathcal{F}^{-1}\left\{\mathcal{F}\{v(x)\}(\omega) * LoG(\omega)\right\}$ derived as follows:

\begin{equation}
\begin{aligned}
     v(x) \odot \mathcal{F}^{-1}\left\{LoG(\omega)\right\}(x) = v(x) \odot \left(-\frac{\sigma x^2}{2\pi} e^{-\frac{(\sigma x)^2}{2}}\right).
\end{aligned}
\end{equation}

The derivation of the 2D LoG kernel is directly applied from that of the 1D LoG kernel.

\subsection{Frequency-level High-pass Filter}
The frequency-level HPF should compress out the low-frequency components in the magnitude spectrum map after applying the pixel-level HPF.
We utilize the ideal HPF that equals $0$ before the predefined cut-off frequency, and $1$ otherwise.
Thus, the output $\mathbf{Z}'$ after applying the frequency-level HPF can be derived as:
\begin{equation}
  \mathbf{Z}'(\omega_1,\omega_2) =
  \begin{cases}
    0       & \quad \text{if } \omega_1^2+\omega_2^2 \leq \omega_c^2\\
    \mathbf{Z}(\omega_1,\omega_2)  & \quad \text{otherwise,}
  \end{cases}
\end{equation}
where $\omega_1\in\{-w/2,\dots,w/2\}$, $\omega_2\in\{-h/2,\dots,h/2\}$, $w_c$ is the predefined cut-off frequency, and $\mathbf{Z}$ is the magnitude spectrum map filtered by the pixel-level HPF.
Then, after applying the frequency-level HPF, all the low-frequency components under $w_c$ are entirely removed, while the high-frequency components over $w_c$ remain.

\section{Experimental Results}
To show the cross-domain performance of the proposed framework, we build three types of experiments as follows: the \textit{cross-category}, \textit{cross-color}, and \textit{cross-GAN performance} evaluations.
First, for \textit{cross-category performance}, we test the model trained with only one category and then with multiple categories.
Second, for \textit{cross-color performance}, color manipulation is considered within the same category and tested with different color variations.
Lastly, for \textit{cross-GAN performance}, the model is trained with only one GAN model and tested with multiple GAN models.

\begin{table}[t!] 
\caption{Cross-category performance with face dataset }\label{table:crossDomain_face}
\resizebox{1.00\linewidth}{!}{%
\begin{tabular}{lcccccc}
\hline
\multicolumn{1}{c}{\multirow{2}{*}{Model}} & \multicolumn{2}{c}{Face} & \multicolumn{2}{c}{Cross-categories} & \multicolumn{2}{c}{All-categories} \\ \cline{2-7} 
                  & Acc. & A.P. & Acc. & A.P. & Acc. & A.P.\\ \hline
Wang (CVPR 2020)& \textbf{99.9} & \textbf{100.0} & 60.9 & 73.6 & 68.7 & 78.9\\
Frank (ICML 2020) & 95.2 & 96.5 & 69.1 & 72.3 & 74.3 & 77.1\\ 
Durall (CVPR 2020) & 86.2 & 93.4 & 62.7 & 53.1 & 67.4 & 61.2\\ 
Ours & 97.0 & 98.1 & \textbf{72.7} & \textbf{76.1} & \textbf{77.6} & \textbf{80.5}\\

\hline
\end{tabular}
}
\end{table}

\begin{table}[t!]
    \caption{Cross-category performance with \textit{horse} category only }
    \label{tab:previous_works}
    \begin{center}
    \resizebox{1.00\linewidth}{!}{%
        \begin{tabular} {llcccccc}
        \hline
        \multicolumn{1}{c}{\multirow{2}{*}{Feat.}} & \multicolumn{1}{c}{\multirow{2}{*}{Classifier}} & \multicolumn{2}{c}{Test-category} & \multicolumn{2}{c}{Cross-categories} & \multicolumn{2}{c}{All-categories}  \\ 
        \cline{3-8} 
        \multicolumn{1}{c}{} & \multicolumn{1}{c}{} & \multicolumn{1}{c}{Acc.} & \multicolumn{1}{c}{A.P.} & \multicolumn{1}{c}{Acc.} & \multicolumn{1}{c}{A.P.}  & \multicolumn{1}{c}{Acc.} & \multicolumn{1}{c}{A.P.}  \\ 
        \hline
        Pixel-based & ResNet-50 & 72.4 & 68.7 & 61.9 & 58.6 &64.5 & 61.1 \\
        Wang (CVPR 2020)& ResNet-50 & 50.5 & 66.6 & 50.0 & 58.3 & 50.2 & 60.4 \\
        Frank (ICML 2020) & ResNet-50 & 93.3 & 89.7 & 73.5 & 68.1 & 78.4 & 73.5\\
        Durall (CVPR 2020) & SVM (rbf) & 88.3 & 83.0 & 62.0 & 59.2 &68.5 & 65.1\\
        Durall (CVPR 2020) & SVM (poly) & 88.8 & 83.9 & 62.0 & 59.1 & 68.7 & 65.3 \\
        Durall (CVPR 2020) & SVM (linear) & 81.1 & 74.1 & 60.2 & 57.0 & 65.4 & 61.3 \\
        Durall (CVPR 2020) & Linear Reg. & 79.9 & 73.2 & 60.5 & 57.0 &65.3 & 61.1 \\
        Ours & ResNet-50 & \textbf{94.8} & \textbf{93.5} & \textbf{73.4} &	\textbf{69.0} & \textbf{78.7} & \textbf{75.2} \\
        \hline
\vspace{-1.5em}
        \end{tabular}
    }
    \end{center}
\end{table}

\subsection{Implementation}\label{sec:implementation}
\noindent\textbf{Hyperparameter. } 
The input size is fixed by $256\times 256$, and Adam optimizer is used to train the classification network with the learning rate of $10^{-4}$.
The updating phase is iterated until $20$ epochs.
In BiHPF, the cutoff frequency $w_c$ is set to $40$, and the variance of LoG filters $\sigma$ is $0.01$. 

\noindent\textbf{Dataset. } 
We train our model with ProGAN~\cite{progan} and test with various images and diverse GAN models, as conducted by Wang~\etal~\cite{adobe}. 
For real images, we use \textit{LSUN}~\cite{lsun} and \textit{FFHQ}~\cite{stylegan}. 
For fake images, we utilize various generated images of different GAN models, including \textit{ProGAN}, 
\textit{StyleGAN}~\cite{stylegan}, \textit{StyleGAN2}~\cite{stylegan2}, \textit{BigGAN}~\cite{biggan}, \textit{CycleGAN}~\cite{cyclegan}, and \textit{StarGAN}~\cite{stargan}.
To validate the effectiveness of BiHPF for cross-domain performance, we utilize various domains including the subject categories, color manipulations, and GAN models.
The generated classes vary by the GAN models, as in Wang~\etal~\cite{adobe}. 
Also, we consider five types of color manipulations including variations of hue, brightness, saturation, gamma, and contrast. 
Since the images have different resolutions, all images are resized into $256\times256$. 

\noindent\textbf{Evaluation metrics. } 
We utilize two evaluation metrics of the average precision score (A.P.) and accuracy (Acc.).
The A.P. is obtained by the alternative measurement used by Wang~\etal~\cite{adobe}, which approximates the area under the precision-recall curve by using a few thresholds.
The Acc. is the proportion of the correctly predicted test images among the entire test images.

\begin{table*}[t!]
\centering
\scriptsize
\caption{Comparison of cross-color performance. }
\label{tab:face_color}
\resizebox{0.93\linewidth}{!}{%
\begin{tabular}{lcc|cccccccccc|cccc}
\hline
\multicolumn{1}{c}{\multirow{2}{*}{Model}} & \multicolumn{2}{c}{Original} & \multicolumn{2}{|c}{Hue} & \multicolumn{2}{c}{Brightness} & \multicolumn{2}{c}{Saturation} & \multicolumn{2}{c}{Gamma} & \multicolumn{2}{c}{Contrast} & \multicolumn{2}{|c}{Mean} & \multicolumn{2}{c}{Min} \\ \cline{2-17} 
 & Acc. & A.P. & Acc. & A.P. & Acc. & A.P. & Acc. & A.P. & Acc. & A.P. & Acc. & A.P. & Acc. & A.P. & Acc. & A.P. \\ \hline
Wang (CVPR 2020) & 99.9 & 100.0 & 73.9 & 81.3 & 61.8 & 74.7 & 74.3 & 84.4 & 70.2 & 83.2 & 66.6 & 79.7 & 69.4 & 80.7 & 61.8 & 74.7 \\
Frank (ICML 2020) & 95.2 & 96.5 & 85.5 & 97.2 & 84.2 & 97.2 & 91.2 & 98.0 & 85.4 & 97.4 & 84.3 & 96.7 & 86.1 & 97.3 & 84.2 & \textbf{96.7} \\
Durall (CVPR 2020) & 86.2 & 93.4 & 86.2 & 81.9 & 85.9 & 81.9 & 86.2 & 81.9 & 85.1 & 80.8 & 85.2 & 81.2 & 85.7 & 81.5 & 85.1 & 80.8 \\
Our & 97.0 & 98.1 & 92.0 & 97.8 & 92.0 & 97.9 & 91.9 & 96.7 & 91.7 & 96.8 & 92.4 & 98.1 & \textbf{92.0} & \textbf{97.5} & \textbf{91.7} & \textbf{96.7} \\
\hline
\vspace{-1.5em}
\end{tabular}
}
\end{table*}

\begin{table*}[t!]
\centering
\scriptsize
\caption{Comparison of cross-model performance. }
\label{tab:confusion_12}
\resizebox{1.00\linewidth}{!}{%
\begin{tabular}{lcccccccccccccc|cccc}
\hline
\multicolumn{1}{c}{\multirow{3}{*}{Model}} & \multicolumn{2}{c}{Training settings} & \multicolumn{14}{c}{Test Models} \\ \cline{2-19} 
& \multirow{2}{*}{Input} & \multirow{2}{*}{\# class} & \multicolumn{2}{c}{ProGAN} & \multicolumn{2}{c}{StyleGAN} & \multicolumn{2}{c}{StyleGAN2} & \multicolumn{2}{c}{BigGAN} & \multicolumn{2}{c}{CycleGAN} & \multicolumn{2}{c}{StarGAN} & \multicolumn{2}{|c}{Mean}& \multicolumn{2}{c}{Min} \\ \cline{4-19} 
 &  &  & Acc. & A.P. & Acc. & A.P. & Acc. & A.P. & Acc. & A.P. & Acc. & A.P.& Acc. & A.P. & Acc. & A.P. & Acc. & A.P. \\ \hline
Wang (CVPR 2020) & Pixel & 1 & 50.4 & 63.8 & 50.4 & 79.3 & 68.2 & 94.7 & 50.2 & 61.3 & 50.0 & 52.9 & 50.0 & 48.2 & 53.2 & 66.7 & 50.0 & 48.2\\
Frank (ICML 2020) & Freq & 1 & 78.9 & 77.9 & 69.4 & 64.8 & 67.4 & 64.0 & 62.3 & 58.6 & 67.4 & 65.4 & 60.5 & 59.5 & 67.7 & 65.0 & 60.5 & 58.6 \\
Durall (CVPR 2020) & Freq & 1 & 85.1 & 79.5 & 59.2 & 55.2 & 70.4 & 63.8 & 57.0 & 53.9 & 66.7 & 61.4 & 99.8 & 99.6 & 73.0 & 68.9 & 57.0 & 53.9 \\
Our & Freq & 1 & 82.5 & 81.4 & 68.0  & 62.8 & 68.8  & 63.6  & 67.0 & 62.5 & 75.5  & 74.2 & 90.1 & 90.1 & \textbf{75.3} & \textbf{72.4} & \textbf{67.0} & \textbf{62.5}\\ \hline
Wang (CVPR 2020) & Pixel  & 2 & 64.6 & 92.7 & 52.8 & 82.8 & 75.7 & 96.6 & 51.6 & 70.5 & 58.6 & 81.5 & 51.2 & 74.3 & 59.1 & \textbf{83.1} & 51.2 & 70.5 \\
Frank (ICML 2020) & Freq & 2 & 85.7 & 81.3 & 73.1 & 68.5 & 75.0 & 70.9 & 76.9 & 70.8 & 86.5 & 80.8 & 85.0 & 77.0 & 80.4 & 74.9 & \textbf{73.1} & 68.5 \\
Durall (CVPR 2020) & Freq & 2 & 79.0 & 73.9 & 63.6 & 58.8 & 67.3 & 62.1 & 69.5 & 62.9  & 65.4 & 60.8 & 99.4 & 99.4 & 74.0 & 69.7 & 63.6 & 58.8 \\
Our & Freq & 2 & 87.4 & 87.4 & 71.6 & 74.1 & 77.0 & 81.1 & 82.6 & 80.6 & 86.0 & 86.6 & 93.8  & 80.8 & \textbf{83.1} & 81.8 & 71.6 & \textbf{74.1}\\
 \hline
Wang (CVPR 2020) & Pixel  & 4 & 91.4 & 99.4 & 63.8 & 91.4 & 76.4 & 97.5 & 52.9 & 73.3 & 72.7 & 88.6 & 63.8 & 90.8 & 70.2 & \textbf{90.2} & 52.9 & 73.3\\
Frank (ICML 2020) & Freq & 4 & 90.3 & 85.2 & 74.5 & 72.0 & 73.1 & 71.4 & 88.7 & 86.0 & 75.5 & 71.2 & 99.5 & 99.5 & 83.6 & 80.9 & 73.1 & 71.2 \\
Durall (CVPR 2020) & Freq & 4 & 81.1 & 74.4 & 54.4 & 52.6 & 66.8 & 62.0 & 60.1& 56.3  &  69.0& 64.0 & 98.1 & 98.1 & 69.7 & 66.6 & 54.4 & 52.6 \\
Our & Freq & 4 & 90.7 & 86.2 & 76.9 & 75.1 & 76.2 & 74.7 & 84.9 & 81.7 & 81.9 & 78.9 & 94.4 & 94.4 & \textbf{84.2} & 81.8 & \textbf{76.2} & \textbf{74.7} \\ \hline
\vspace{-1.5em}
\end{tabular}
}
\end{table*}

\subsection{Cross-domain Performance Evaluations}

\begin{figure*}[t]
\centering
        \subfigure[Low-pass filter]{\includegraphics[width=0.49\linewidth]{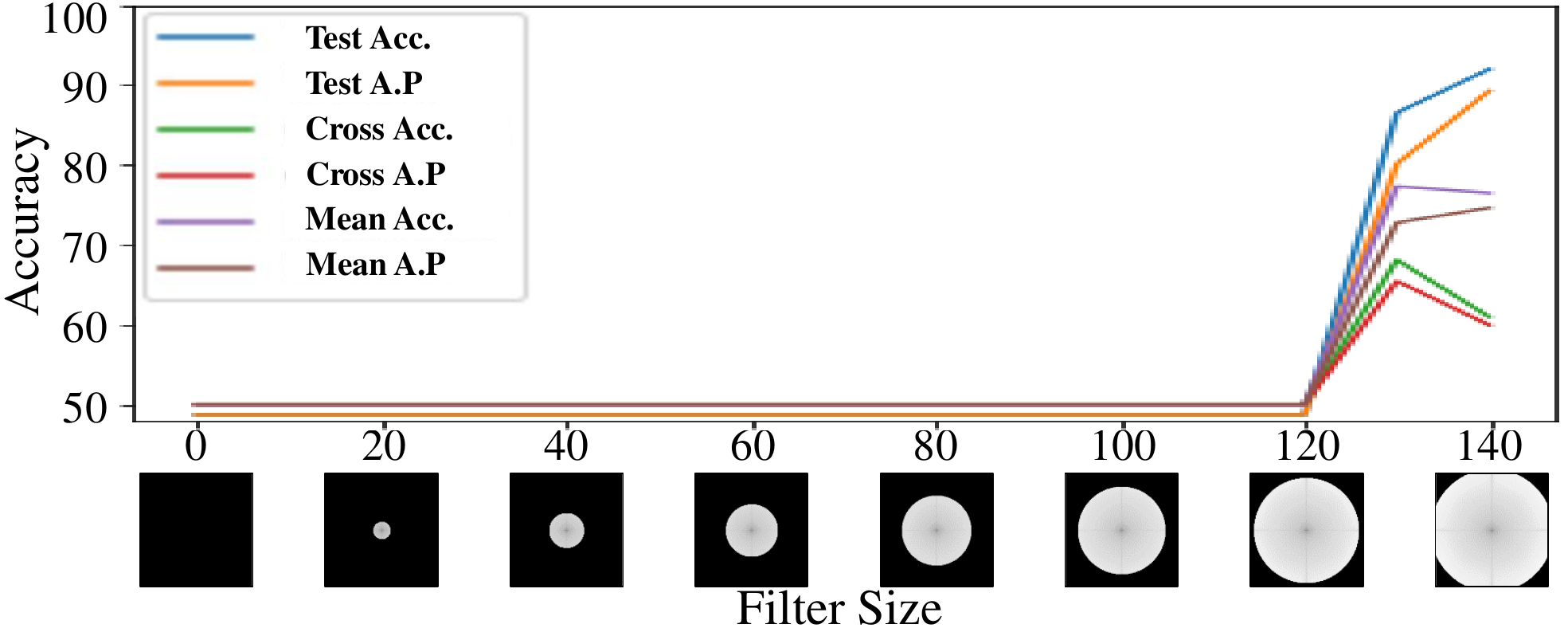}}
        \subfigure[High-pass filter]{\includegraphics[width=0.49\linewidth]{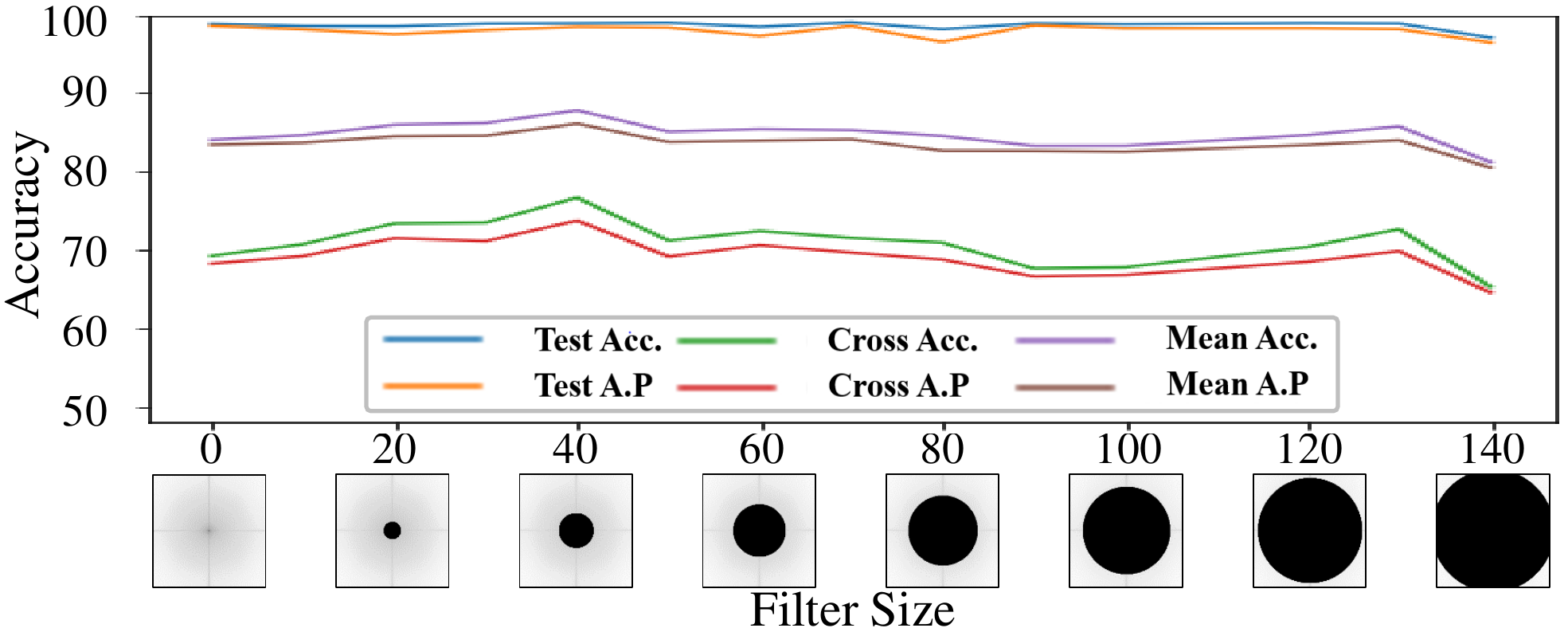}}
    \caption{Ablation test with various $w_c$}
    \vspace{-1.5em}
    \label{fig:masking}
\end{figure*}

\noindent\textbf{Cross-category Performance. } We perform two experiments to show the cross-category performance.
First, we train the detection model with the \textit{face} category of fake images generated by \textit{ProGAN} and evaluate it with the remaining 20 categories. 
The number of training images is $20,000$ and each category of the test set contains $400$ images. The number of real images is always the same as the number of fake images in each dataset.
The experimental results are given in Table~\ref{table:crossDomain_face}.
The performance in the columns of \textit{Face} is obtained only by the test set of \textit{face} category, while the columns of \textit{Cross-categories} and \textit{All-categories} show the average performance of the remaining four categories and all categories in the testset, respectively.
From the results, we can confirm that the proposed algorithm shows state-of-the-art performance when the same category is considered in the test phase.
Interestingly, although trained only with the \textit{face} category, BiHPF maintains high performance when tested with other categories.

Second, the \textit{horse} category of \textit{ProGAN} is employed as the training data, and the testset contains four categories generated by the same GAN model (ProGAN) as follows: \textit{horse}, \textit{car}, \textit{cat}, and \textit{church}. 
To further evaluate the model with more datasets compared to Wang~\etal~\cite{adobe}, we obtain four additional categories from ProGAN, each containing 1,000 real and 1,000 fake images.
In Table~\ref{tab:previous_works}, the three columns of \textit{Test-category}, \textit{Cross-categories}, and \textit{All-categories} represent the test results in the corresponding category, the other categories, and all four categories, respectively.
The \textit{pixel-based detector} utilizes the original images for the fake image detection, while its training method and architecture are equivalent to the proposed framework.
From the results, our approach successfully improves the cross-category performance compared to the previous studies.
In particular, our model outperforms others in generalized detection based on the simple operations of BiHPF.

\noindent\textbf{Cross-color Performance. } We conduct experiments to validate the robustness of the detection models in color manipulations.
If images are partially manipulated with variations in hue, brightness, saturation, gamma, and contrast, the image artifacts contain different characteristics compared to those in the entirely synthesized images. 
The hue factor is the amount of shift in hue channel by 0.2, while brightness, saturation, gamma, and contrast are adjusted by 1.3, respectively.
Table~\ref{tab:face_color} indicates the variance in detecting performance when images are manipulated and the characteristics of the artifacts have changed.
Impressively, our model is more robust and resistant to diverse manipulations compared to others. 
Especially, Wang~\etal~\cite{adobe} show a decline in performance since it uses RGB rather than grayscale. 

\noindent\textbf{Cross-model Performance. } 
In addition to the superior cross-category and cross-color performances, we discover that our model also works generally across various GAN models.
To show the property, we use the experimental settings of Wang~\etal~\cite{adobe}, which are designed to show the generality of the fake image detector trained by a specific GAN model to identify other GAN models.
We use three types of training settings, which contain 1-class, 2-class, and 4-class settings. 
As represented in Table~\ref{tab:confusion_12}, our proposed approach shows improved performance on the various GAN models, even though the detection algorithm is trained only with the fake images generated by ProGAN.
From the result, we can validate that the proposed approach can be expanded to consider the generality of the artifacts across various GAN models.
The experimental settings are provided in Appendix D in detail.

\begin{table}[t!]
    \caption{Component-wise ablation tests. }
    \vspace{-1.0em}
    \label{tab:Ablation}
    \begin{center}
    \resizebox{1.00\linewidth}{!}{%
        \begin{tabular} {llp{2mm}p{2mm}cccc}
        \hline
        \multicolumn{1}{c}{\multirow{2}{*}{Feat.}} & \multicolumn{1}{c}{\multirow{2}{*}{Classifier}} & \multirow{2}{*}{$\mathcal{L}$} & \multirow{2}{*}{$\mathcal{F}$} & \multicolumn{2}{c}{Test-category} & \multicolumn{2}{c}{Cross-categories} \\ 
        \cline{5-8}
         \multicolumn{1}{c}{} & \multicolumn{1}{c}{} &  &  & \multicolumn{1}{c}{Acc.} & \multicolumn{1}{c}{A.P.} & \multicolumn{1}{c}{Acc.} & \multicolumn{1}{c}{A.P.}\\ \hline
        Durall (CVPR2020) & SVM (rbf) &  & & 92.8 & 88.4 & 70.2 & 66.5 \\
        Durall (CVPR2020) & SVM (linear) &  &  & 86.0 & 78.9 & 68.0 & 63.4  \\
        Durall (CVPR2020) & Logistic Reg. &  &  & 83.9 & 76.7 & 67.7 & 63.1   \\
        Ours & ResNet-50 &  &  & 98.4 & 97.9 & 68.7 & 67.7  \\ \hline
        Durall (CVPR2020) & SVM (rbf) & \checkmark & & 91.1 & 86.1 & 70.4 & 66.4   \\
        Durall (CVPR2020) & SVM (linear) & \checkmark & & 84.3 & 76.4 & 68.2 & 63.5   \\
        Durall (CVPR2020) & Logistic Reg. & \checkmark & & 82.0 & 74.4 & 67.6 & 62.9  \\
        Ours & ResNet-50 & \checkmark & & 98.9 & \textbf{98.6} & 73.1 & 72.5  \\ 
        \hline
        Durall et al. (CVPR 2020) & SVM (rbf) &  & \checkmark & 92.6 & 88.2 & 71.7 & 67.6  \\
        Durall et al. (CVPR 2020) & SVM (linear) &  & \checkmark & 86.3 & 79.0 & 70.0 & 64.9   \\
        Durall et al. (CVPR 2020) & Logistic Reg. & & \checkmark & 84.7 & 77.5 & 71.2 & 65.8  \\
        Ours & ResNet-50 & & \checkmark & 99.0 & 98.2 & 77.7 & 75.4  \\ \hline
        Durall et al. (CVPR 2020) & SVM (rbf) & \checkmark & \checkmark & 91.4 & 86.3 & 69.6 & 64.9 \\
        Durall et al. (CVPR 2020) & SVM (linear) & \checkmark & \checkmark & 84.8 & 76.9 & 69.1 & 64.2 \\
        Durall et al. (CVPR 2020) & Logistic Reg. & \checkmark & \checkmark & 83.3 & 75.6 & 70.1 &  64.9 \\
        Ours & ResNet-50 & \checkmark & \checkmark & \textbf{98.9} & 98.5 & \textbf{79.8} & \textbf{77.7} \\ 
        \hline
        \end{tabular}
    }
    \end{center}
\vspace{-1.5em}
\end{table}

\subsection{Ablation Study}
We validate the performance of each component and hyperparameter of BiHPF through experiments for ablation study. 
In this section, we employ the \textit{horse} dataset of \textit{StyleGAN2} for training to show the model's stable performance even when trained with a different dataset other than \textit{ProGAN}. 
Also, we conduct cross-category experiments to further validate the model's performance. The additional ablation tests are provided in Appendix E. 

\subsubsection{Validity of BiHPF. }
To show that the effect of BiHPF is not limited to our framework, we employ BiHPF to other algorithms for fake image detection~\cite{watch_cvpr20}.
In this experiment, we also compare the effect of the pixel-level HPF ($\mathcal{L}$) and the frequency-level HPF ($\mathcal{F}$) by removing either one or both of them from our mechanism.
Table~\ref{tab:Ablation} shows the results that our mechanism can improve the cross-domain performance compared to the other algorithms for fake image detection.
Thus, by using the proposed mechanism as pre-processing, every single one of the previous studies can preserve its performance when the given subject is out of the training setting.
In addition, when utilizing both of the HPFs in BiHPF, our framework shows superior cross-domain performance than the framework where one or both of the HPFs are missing.
The frequency-level HPF shows greater performance improvement than the pixel-level HPF, signifying the importance of dismissing the low-frequency components to emphasize the effect of the artifacts. 

More specifically, the artifacts in the foreground or center of the image contain domain-specific features, whereas the artifacts in the background or edge of the image contain general information.
Thus, we can confirm that employing the LoG filter is effective at improving the cross-domain performance.
Since Durall~\etal~\cite{watch_cvpr20} does not use the spacial information, it shows poor performance in the cross-domain test.
\begin{table}[t!]
\centering
\caption{Ablation test with various $\sigma$. }
\vspace{-1.0em}
    \label{tab:sigma}
    \begin{center}
    \resizebox{0.8\linewidth}{!}{%
        \begin{tabular} {ccccccc}
        \hline
        \multicolumn{1}{c}{\multirow{2}{*}{$\sigma$}} & \multicolumn{2}{c}{Test-category}                  & \multicolumn{2}{c}{Cross-categories}                 & \multicolumn{2}{c}{All-categories}                         \\ \cline{2-7} 
        \multicolumn{1}{c}{}                       & \multicolumn{1}{c}{Acc.} & \multicolumn{1}{c}{A.P.} & \multicolumn{1}{c}{Acc.} & \multicolumn{1}{c}{A.P.} & \multicolumn{1}{c}{Acc.} & \multicolumn{1}{c}{A.P.} \\ \hline
        0.0001 & \textbf{99.0} & 98.4 & 79.3 & 76.8 & 84.2 & 82.2 \\
        0.001 & 98.2 & 97.0 & 78.0 & 75.7 & 83.1 & 81.0 \\
        0.01 & 98.9 & \textbf{98.5} & \textbf{79.8} & \textbf{77.7} & \textbf{84.6} & \textbf{82.9} \\
        0.1 & 98.8 & 97.8 & 76.5 & 74.0 & 82.1 & 79.9 \\
        0.5 & 98.9 & 98.4 & 74.9 & 73.7 & 80.9 & 79.9 \\
        \hline
        \end{tabular}
    }
    \end{center}
    \vspace{-1.0em}
\end{table}

\subsubsection{Hyperparameters of BiHPF. }
The frequency-level and pixel-level HPFs of BiHPF are respectively controlled by only one hyperparameter for each: the cut-off frequency ($w_c$) and the variance of LoG ($\sigma^2$).
First, to show the sensitivity of the performance with $\sigma$ values, we perform additional experiments by changing the value of $\sigma$ as represented in Table~\ref{tab:sigma}.
The performance is evaluated by the validation set different from the test set, and the best performance overall is when $\sigma=0.01$.
Interestingly, while the test domain performance fluctuates with the various values of $\sigma$, the cross-domain performance consistently declines as $\sigma$ moves far from the peak of $\sigma=0.01$.
Thus, we can reconfirm that the value of $\sigma$ influences the effect of the artifacts in fake images.

\begin{table}[t!]
    \caption{Performance comparison between colored images and grayscale images as inputs. } 
    \vspace{-1.0em}
    \label{tab:RGB_vs_Gray}
    \begin{center}
    \resizebox{1.00\linewidth}{!}{%
    \begin{tabular} {llcccccc}
    \hline
    \multicolumn{1}{c}{\multirow{2}{*}{}}                                                  & \multicolumn{1}{c}{\multirow{2}{*}{Scale}} & \multicolumn{2}{c}{Test-category}                           & \multicolumn{2}{c}{Cross-categories}                          & \multicolumn{2}{c}{All-categories}                                  \\ \cline{3-8} 
    \multicolumn{1}{c}{}                                                                   & \multicolumn{1}{c}{}                       & \multicolumn{1}{c}{Acc.}     & \multicolumn{1}{c}{A.P.}      & \multicolumn{1}{c}{Acc.}     & \multicolumn{1}{c}{A.P.}      & \multicolumn{1}{c}{Acc.}     & \multicolumn{1}{c}{A.P.}      \\ \hline
    \multirow{2}{*}{\begin{tabular}[c]{@{}l@{}}Pixel-level\\ (image)\end{tabular}} & RGB & \textbf{99.6} & \textbf{99.5} & 54.7 & 54.7 & 65.9 & 65.9  \\
     & Gray & 99.5 & 99.2 & 50.4 & 50.4 & 62.7 & 62.6 \\ 
     \hline
    \multirow{2}{*}{\begin{tabular}[c]{@{}l@{}}Frequency-level\\ (magnitude)\end{tabular}} & RGB                                        & 99.6 & 99.2 & 57.7 & 57.7 & 68.2 & 68.1 \\
     & Gray & 98.4 & 97.87 & \textbf{68.4} & \textbf{67.7} & \textbf{76.1} & \textbf{75.3} \\ 
        \hline
        \end{tabular}
    }
    \end{center}
\vspace{-1.5em}
\end{table}

Second, we validate the necessity of the frequency-level HPF by replacing it with the frequency-level low-pass filter.
As shown in Fig.~\ref{fig:masking}~(a), the performance declines dramatically when the high-frequency components are removed by the frequency-level low-pass filter, which validates the importance of the high-frequency components for fake image detection.
In contrast, as shown in Fig.~\ref{fig:masking}~(b), the performance maintains even with the various values of $w_c$.
Thus, the performance is robust to $w_c$, which indicates that the fake images can be distinguished even with limited high-frequency components in the Fourier domain.

\subsubsection{Comparison of RGB and Gray Images. }
The color information contains class-specific characteristics that reduce the effect of the artifacts, which leads to a decline in cross-domain performance.
Table~\ref{tab:RGB_vs_Gray} indicates the comparison between the colored and grayscale images. The color information in the pixel-based approach leads to improved performance in the test domain than in the cross-domain, which also happens in the frequency-based approach. 
Thus, we can confirm that the color information hinders amplifying the effect of the artifacts. 
The experiments are conducted based on Resnet-50~\cite{resnet}.

\section{Conclusion}
We present a novel mechanism called BiHPF for robust detection of synthesized images across various categories, color manipulations, and GAN models.
Since the previous state-of-the-art models heavily depend on the training settings, their performance can decline when tested with unseen data during the training phase.
In contrast to those models, our model achieves robust generalization with state-of-the-art performance, when tested even with unseen data outside of the training settings. 
Our new model has strong practical implications in three major aspects: first, our model achieves superior performance in both facial and non-facial categories, unlike most other models with limited detection performance in the face category only; 
second, our new model shows the outstanding cross-domain performance when tested with various categories; 
finally, we develop a simple but robust model adversarially extracting the frequency-level artifacts, which are successfully utilized for the detection of synthesized images of GAN models.
We expect our new framework to be utilized in the prevention and detection of malicious abuse of synthesized images to protect our society. 
\footnote{https://github.com/SamsungSDS-Team9/BiHPF}

\bibliographystyle{ieee_fullname}
\bibliography{main}

\end{document}